\documentclass[letterpaper, 10 pt, conference]{ieeeconf}  

\IEEEoverridecommandlockouts                              

\usepackage{times}

\usepackage{multicol}
\usepackage[dvipsnames]{xcolor}
\usepackage{graphicx}
\usepackage{hyperref}
\usepackage{amsmath}
\usepackage{cleveref}
\usepackage[ruled,vlined,linesnumbered]{algorithm2e}
\usepackage[font=footnotesize,labelfont=bf]{caption}
\usepackage{subcaption}
\usepackage{bbm}
\usepackage{float}
\usepackage{color, soul} 
\usepackage{wrapfig,lipsum,booktabs}

\usepackage{array}
\usepackage[utf8]{inputenc}
\usepackage{amsmath, amssymb}
\usepackage{booktabs}
\usepackage{dsfont}
\usepackage{diagbox}
\usepackage{multirow}

\usepackage{amsmath,amsfonts,bm}

\def\eqref#1{equation~\ref{#1}}

\def\1{\bm{1}}

\DeclareMathAlphabet{\mathsfit}{\encodingdefault}{\sfdefault}{m}{sl}
\SetMathAlphabet{\mathsfit}{bold}{\encodingdefault}{\sfdefault}{bx}{n}

\usepackage{algorithmic}
\usepackage{graphicx}   
\usepackage{capt-of}
\usepackage{etoolbox}    
\usepackage{stfloats}   

\title{\LARGE \bf
Multimodal Diffusion Forcing for Forceful Manipulation
}

\usepackage{cuted}     
\newcommand{\TopTeaser}{%
  \begin{strip}
    \centering
    \vspace{-48pt}
    \includegraphics[width=\linewidth]{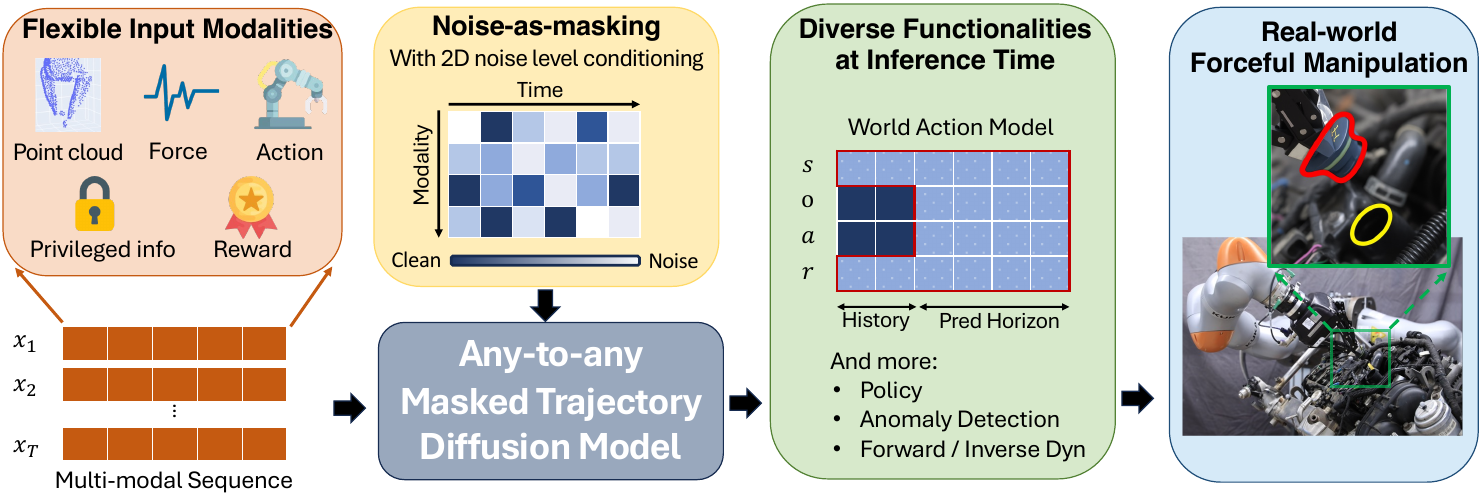}
    \captionof{figure}{We propose \emph{Multimodal Diffusion Forcing}, a unified model that captures the interplay between modalities over time through partially masked training. At inference time, the model not only offers flexibility by allowing different input modalities, adjustable history lengths and prediction horizons, it also provides diverse functionalities — serving as a policy, world action model, dynamics model, and anomaly detector.
    }
    \label{fig:teaser}
  \end{strip}%
}

\makeatletter
\let\orig@maketitle\maketitle
\renewcommand{\maketitle}{%
  {\let\newpage\relax\orig@maketitle}

  \TopTeaser
}
\makeatother

\author{Zixuan Huang, Huaidian Hou, Dmitry Berenson \\
University of Michigan, Ann Arbor
  \thanks{This work was supported in part by the Office of Naval Research under grants N00014-24-1-2036 and N6833525C0329 and NSF grants IIS-2113401 and IIS-2220876. Robotics Department, 
        University of Michigan, Ann Arbor, MI, USA, contact
        {\tt\small zixuanh@umich.edu}.}
}

\begin{document}

\maketitle
\thispagestyle{empty}
\pagestyle{empty}

\begin{abstract}
Given a dataset of expert trajectories, standard imitation learning approaches typically learn a direct mapping from observations (e.g., RGB images) to actions. However, such methods often overlook the rich interplay between different modalities, i.e., sensory inputs, actions, and rewards — which is crucial for modeling robot behavior and understanding task outcomes. In this work, we propose \emph{Multimodal Diffusion Forcing}, a unified framework for learning from multimodal robot trajectories that extends beyond action generation. 
Rather than modeling a fixed distribution, MDF applies random partial masking and trains a diffusion model to reconstruct the trajectory. This training objective encourages the model to learn temporal and cross-modal dependencies, such as predicting the effects of actions on force signals or inferring states from partial observations. 
We evaluate MDF on contact-rich, forceful manipulation tasks in simulated and real-world environments. Our results show that MDF not only delivers versatile functionalities, but also achieves strong performance, and robustness under noisy observations. More visualizations can be found on our website \href{https://unified-df.github.io/}{https://unified-df.github.io}
\end{abstract}



\section{Introduction}
Humans naturally integrate visual, audio, tactile, and proprioceptive signals to understand and interact with the physical world. For example, when inserting a key into a lock or tightening a bolt, we adjust our motion based on visual alignment and subtle resistance felt through touch. Similarly, robots performing contact-rich tasks must reason over diverse sensory inputs to perceive object states, predict outcomes, and react appropriately.

Despite the richness of multimodal sensory data in robotic systems, most existing learning methods~\cite{du2022play, chen2022visuo, liu2024maniwav,lin2024learning} focus on direct mappings from observations to actions. These approaches often overlook the complex interplay between modalities. Furthermore, existing approaches typically assume a fixed set of input modalities and lack robustness to partial or corrupted observation at inference time.

In this work, we propose \emph{Multimodal Diffusion Forcing} (MDF), a unified framework for learning the joint distribution of multimodal robot trajectory. Unlike standard diffusion models that use a single, global noise level, MDF is trained using a 2D Time-Modality Noise Level Matrix. This unique training scheme endows MDF with the following properties:

\noindent\textbf{Capturing cross-modal correlations over time.} Our masked training strategy randomly corrupts input modalities and requires the model to recover them from the remaining context. This encourages the model to learn temporal dependencies both within and across modalities.

\noindent\textbf{Flexibility at training and inference time.} The model is trained to condition on arbitrary subsets of modalities and predict the rest, all controlled via the 2D noise level matrix. During training, we can include privileged modalities such as full point clouds, even if they are not available at test time. This form of privileged learning has been shown to improve robustness by encouraging the model to infer privileged information from partial observations~\cite{kumar2021rma, akinola2025tacsl}. At inference time, a single MDF model can be flexibly deployed for a range of downstream tasks, including policy, dynamics modeling, and fine-grained anomaly detection.

\noindent\textbf{Robustness to noisy inputs.} Since MDF is trained with a \textit{continuous} scale of corruption level, it is robust to a wide range of noisy or missing observations at test time compared to models trained with binary masking.

We evaluate MDF in five contact-rich forceful manipulation tasks in simulation and the real world. All tasks require high-precision prediction and multimodal reasoning. We show that the performance of MDF is on par with specialized models and may surpass them when noisy observations are present. We also demonstrate the wide range of test-time capabilities of MDF.


\section{Related work}
\label{sec:related_work}

\subsection{Diffusion Models for Robotics}
Diffusion models have been widely adopted in robotics for tasks such as policy learning~\cite{chi2023diffusion, chen2023playfusion, xue2025reactive, ze20243d}, forward dynamics modeling~\cite{ding2024diffusion, rigter2023world}, subgoal prediction~\cite{ajay2022conditional, huang2024subgoal, huang2024implicit, hsu2024spot}, and joint state-action modeling~\cite{janner2022planning}. They have also been used for anomaly detection by estimating the log-likelihood via a variational bound~\cite{jianganomalies}. Most prior models are trained for a single task. In contrast, our model's modality- and timestep-specific masking allows zero-shot adaptation to a variety of downstream tasks, including action generation, dynamics prediction, state estimation, and anomaly detection.

Recently, UWM~\cite{zhu2025unified} and UVA~\cite{li2025unified} introduced similar unified frameworks for behavior learning. While they primarily focus on video and action, our approach generalizes to multimodal sequences incorporating point clouds and force. Moreover, our method is more robust to sensory noise and offers greater test-time flexibility with configurable history lengths and input modalities. Due to the demanding compute requirement of UVA, we only compare with UWM.

\subsection{Learning from Multimodal Data}
There has been a growing interest in integrating multimodal information into robotic systems, including vision~\cite{nair2022r3m, ze20243d}, language~\cite{black2410pi0, jiang2022vima}, audio~\cite{du2022play, li2022see}, force~\cite{noseworthy2025forge, kang2025robotic, wu2024tacdiffusion}, and tactile signals~\cite{chen2022visuo, cui2020self, akinola2025tacsl, collins2024forcesight}. A common paradigm is to train policies that directly fuse multi-sensor inputs into actions. More recently, several works have explored jointly predicting both future actions and observations, such as images~\cite{guo2024prediction}, force~\cite{collins2024forcesight}, or tactile signals~\cite{heng2025vitacformer}, and have shown that this improves policy performance. However, these approaches typically rely on fixed input–output structures and assume access to a complete and consistent set of modalities.
In contrast, MDF generalizes across varying input–output structures and supports multiple functionalities beyond action generation. Moreover, its noise-as-masking training paradigm provides richer supervision signals beyond dynamics learning alone.

\subsection{Masked Training for Robotics }
Recently, researchers attempted to apply masked training to robotics~\cite{wu2023masked, carroll2022uni, radosavovic2024humanoid, liu2022masked, li2025unified, zhu2025unified}, demonstrating the benefits of multimodal joint learning and the flexibility it offers for various downstream tasks. However, existing methods often rely on the low-dimensional state and struggle to handle high-dimensional observation~\cite{wu2024tacdiffusion, carroll2022uni}. Moreover, many approaches adopt a binary masking scheme~\cite{wu2024tacdiffusion, carroll2022uni, radosavovic2024humanoid, liu2022masked, li2025unified}. This limits the models to only reasoning about fully clean or completely missing data, but not anything in between. In contrast, we built on the idea of noise-as-masking~\cite{chen2024diffusion} and train MDF with a continuous masking theme. This design enables fine-grained partial masking across time and modality, allowing the model to reason over partially corrupted observations robustly.



\begin{figure*}[!t]
    \centering
    \includegraphics[width=1\linewidth]{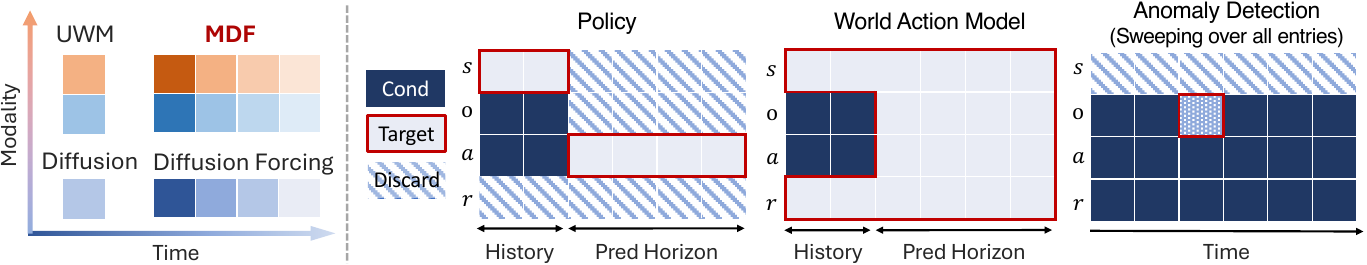}
    \caption{Typical diffusion models use a scalar noise level to control the denoising process. Diffusion Forcing~\cite{chen2024diffusion} proposes a time-varying noise vector to sample video sequences autoregressively. We further generalize this framework to the multimodal setting by introducing a time–modality varying noise matrix. This design enables versatile functionalities at test time such as policy, world action model, dynamics model, and fine-grained anomaly detector. }
    \label{fig:noise-schedule-combined}
    \vspace{-5mm}
\end{figure*}

\section{Preliminaries}

\subsection{Diffusion Models}
Diffusion models gradually transforms data into noise through a forward Markov process, and then learns to reverse this process to reconstruct the original data distribution. The forward process \( q(x^k \mid x^{k-1}) \) adds Gaussian noise at each diffusion step \( k \), defined as:
\[
q(x^k \mid x^{k-1}) = \mathcal{N}\left(x^k \mid \sqrt{1 - \beta_k} \, x^{k-1}, \beta_k \mathbf{I} \right),  k = 1, \ldots, K
\]
where \( \beta_k \) controls the noise variance at each step. 
The overall generative objective is to maximize the log-likelihood of reverse diffusion process under the learned model \( p_\theta \), which is intractable and thus optimized via a variational lower bound:
\begin{align}
 \mathbb{E}_{q(x^0)}[\log p_\theta(x^0)] 
 &\geq \mathbb{E}_{q(x^{0:K})} \left[ \log \frac{p_\theta(x^{0:K})}{q(x^{1:K} \mid x^0)} \right] \nonumber 
\end{align}
As shown in DDPM~\cite{ho2020denoising}, the distribution of the reverse process can be represented as $p_\theta(x^{k-1} \mid x^k) = \mathcal{N}\left(x^{k-1}; \boldsymbol{\mu}(x^k, k), \gamma_k \mathbf{I} \right)$, where $\gamma_k$ is a fixed constant depending on $k$. Then, DPPM~\cite{ho2020denoising} proposed to reparametrize the mean $\mu$ with noise prediction according to $\epsilon=(\sqrt{1-\bar{\alpha}_k})^{-1}x^k - \sqrt{\bar{\alpha}_k}\mu$, which leads to the following objective:
\begin{equation}
\label{eq:diffusion_train_obj}
    \mathcal{L}(\theta) = \mathbb{E}_{k, \mathbf{x}^0, \boldsymbol{\epsilon}} \left[ \left\| \boldsymbol{\epsilon}^k - \boldsymbol{\epsilon}_\theta(\mathbf{x}^k, k) \right\|^2 \right]
\end{equation}

\subsection{Problem Setup}
\label{sec:problem_setup}

We consider the problem of modeling multimodal robot trajectories from an offline dataset of expert demonstrations. We represent a trajectory as $\tau = \{ \mathbf{x}_1, \ldots, \mathbf{x}_T \},$
where each timestep \( \mathbf{x}_t = \{ \mathbf{x}_{t,1}, \ldots, \mathbf{x}_{t,m} \} \) consists of a set of \( M \) modality-specific features. Here, the notion of "modality" is general: it encompasses not only multimodal observations \( o_t \in \mathcal{O} \) (e.g., images, point clouds, force signals), but also actions \( a_t \in \mathcal{A} \), rewards \( r_t \in \mathcal{R} \), and privileged states \( s_t \in \mathcal{S} \) (e.g., object pose, velocity, or full-scene reconstructions) that are typically unavailable at test time. This abstraction allows us to build a unified model capable of handling heterogeneous data types within a single framework.

Our objective is to learn a unified generative model over multimodal robot trajectories that captures the complex interplays of different modalities and can be used for various downstream tasks.

\section{Method}
In section~\ref{sec:multimodal_diffusion_forcing}, we introduce the generalized multimodal diffusion forcing training scheme and the architecture. In section~\ref{method:inference}, we present the flexible inference-time capability of MDF. In section~\ref{method:training}, we summarize the key training and implementation details.

\subsection{Generalized Masked Training with Multi-modal Diffusion Forcing}
\label{sec:multimodal_diffusion_forcing}

\paragraph{Noise as masking} Masked training approaches~\cite{kenton2019bert, he2022masked} usually train models to predict missing parts of the input with a binary mask. Diffusion models, which learn to iteratively denoise Gaussian-corrupted inputs, can be viewed as a continuous masking scheme where noise serves as partial masking~\cite{chen2024diffusion}. A noise level of zero denotes an unmasked token, whereas a maximal noise level corresponds to complete masking. The partial masking aligns more with natural corruption in robotics (i.e. noisy state estimation, partial occlusion). 

However, standard diffusion models typically apply a scalar noise level uniformly across all data, such as a full image~\cite{ho2020denoising}, video~\cite{ho2022video} or trajectory~\cite{janner2022planning}. This design introduces two limitations. First, during training, the model underutilizes the multimodal supervision by learning only a fixed global corruption pattern. Second, at test time, it lacks the flexibility to selectively condition on arbitrary subsets of modalities and timesteps, which is critical for partially observed or corrupted sequences.

To address these challenges, we extend Diffusion Forcing~\cite{chen2024diffusion} to the multimodal setting by introducing a 2D \textit{Time-Modality Noise Level Matrix} $\mathbf{K}\in \{0,\ldots,K\}^{T \times M}$, where $K$ is the number of diffusion steps, $M$ is the number of modalities, and $T$ is length of the trajectory (Fig.~\ref{fig:noise-schedule-combined}.
$k_{t,m}$ specifies the noise level applied to modality $m$ at timestep $t$. During training, the Gaussian noise $\epsilon$ and noise levels $k$ are sampled independently across modalities and timesteps, allowing each part of the multimodal sequence to be partially corrupted to a different extent. Given a noise level matrix $\mathbf{K}$, the forward diffusion process can be written as follows:
\[
\mathbf{x}_{t, m}^{k_{t,m}} = \sqrt{\bar{\alpha}_{k_{t,m}}} \, \mathbf{x}_{t, m}^0 + \sqrt{1 - \bar{\alpha}_{k_{t, m}}} \, \boldsymbol{\epsilon}_{t, m}, \quad \boldsymbol{\epsilon}_{t, m} \sim \mathcal{N}(\mathbf{0}, \mathbf{I})
\]
Here, \( \mathbf{x}_{t, m}^{k_{t,m}} \) denotes the noised version of the original clean feature \( \mathbf{x}_{t,m}^0 \) at timestep \( t \) for modality \( m \), corrupted according to its assigned noise level \( k_{t,m} \). Thus, the multimodal diffusion forcing model can be parameterized by \( \epsilon_\theta\left( \tau^{\mathbf{K}}, \mathbf{K} \right)\), which takes the entire multimodal sequence and the 2D noise level matrix as input. The model is trained with the standard DDPM objective~\cite{ho2020denoising}:
\[
\mathcal{L} = \mathbb{E}_{\mathcal{X}, \mathbf{K}, \boldsymbol{\epsilon}} \left[ \sum_{t=1}^{T} \sum_{m=1}^{M} \left\| \epsilon_{t,m} - \epsilon_\theta\left(\tau^{\mathbf{K}}, \mathbf{K} \right)_{t,m} \right\|^2 \right],
\]

\begin{figure*}
    \centering
    \includegraphics[width=0.95\linewidth]{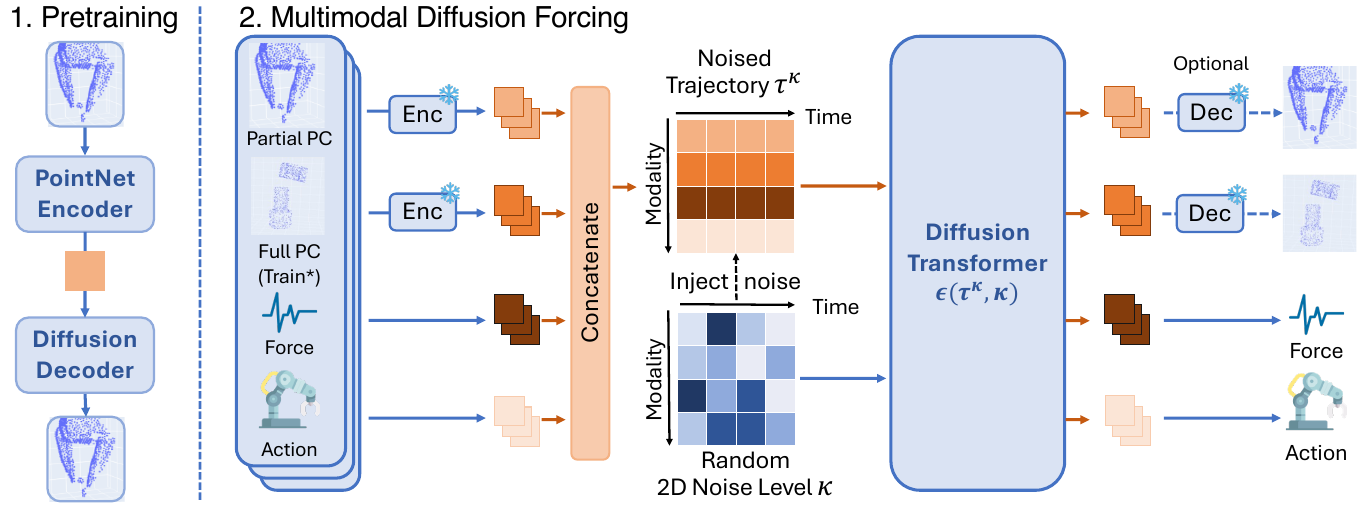}
    \caption{\textbf{Pretraining}: MDF learns a diffusion-based autoencoder to compress point clouds into compact embeddings. \textbf{Multimodal masked training}: MDF processes six modalities: processes six modalities: partial point cloud, full point cloud (training only), force, action, reward, and proprioception; the reward and proprioception are omitted in the figure for clarity. The point clouds are tokenized using the pretrained PointNet encoder (frozen during MDF training). Data from all modalities are then concatenated and corrupted with noise according to a randomly sampled 2D noise level matrix. The diffusion transformer is trained to denoise this corrupted input, learning temporal and cross-modal dependencies.}
    \label{fig:system}
    \vspace{-5mm}
\end{figure*}

In this work, we consider six modalities (Fig.~\ref{fig:system}): partial point cloud, full point cloud (training-time only), force, action, proprioception, and reward. Although full object point clouds are only available in simulation, we include it to encourage the model to implicitly reason about the objects interaction given partial point cloud. This can be seen as a form of occlusion reasoning or privileged learning~\cite{kumar2021rma}. As shown in the experiments, we find it to be critical to the performance of our method.

\paragraph{Architecture}
Our multimodal diffusion forcing model is implemented as a bi-level diffusion framework (Fig.~\ref{fig:system}), consisting of diffusion-based point cloud autoencoders and a latent diffusion transformer.

Given a multimodal trajectory, we first transform the raw features of each modality into compact vector embeddings suitable for sequence modeling.
While prior works~\cite{zhu2025unified, li2025unified} primarily use sequences of images as observation, we focus on point clouds, as they capture rich geometric information that is critical for manipulation tasks. However, unlike images that can be efficiently encoded/decoded using a VQ-VAE~\cite{van2017neural}, point clouds are high-dimensional and unordered, making them considerably more challenging to model. To obtain meaningful embeddings efficiently, we train a diffusion-based point cloud autoencoder~\cite{luo2021diffusion} composed of a lightweight PointNet encoder and an expressive diffusion decoder. The autoencoder is trained to map point clouds into low-dimensional latent embeddings and reconstructs them through iterative denoising.

Then, we adopt a latent diffusion transformer~\cite{chen2024diffusion} for sequence modeling. At each timestep, we concatenate encoded feature vectors of each modality together with their noise-level embeddings, producing a fused multimodal feature vector. The resulting sequence of fused vectors is then fed into a latent diffusion transformer, which models the bidirectional temporal dependencies and captures cross-modal interactions over time. Importantly, the multimodal diffusion model operates entirely in the latent space, and the expensive point cloud decoding step can be skipped at test time.

\subsection{Flexible Inference-time Capabilities}
\label{method:inference}
During inference, we can query the trajectory diffusion model $\epsilon_\theta$ as an arbitrary conditional distribution by configuring the noise level matrix $\mathbf{K}$. As illustrated in Fig.~\ref{fig:noise-schedule-combined}, the matrix is partitioned into three types of blocks. \textbf{Condition blocks} are assigned near-zero noise levels throughout the diffusion process, ensuring that their information is preserved. \textbf{Target blocks} are initialized with Gaussian noise and assigned a high noise level. This noise level is then gradually denoised to zero according to a predefined noise schedule. Although different denoising schedules can be specified for each entry~\cite{chen2024diffusion}, we find that denoising the full sequence with a uniform timestep across target modalities is more efficient than autoregressive schemes. Finally, \textbf{discard blocks} are assigned the maximum noise level and remain noisy throughout the denoising process. 

With these three types of blocks, we can flexibly configure the sampling distribution to support diverse functionalities. For instance, the model can function as a \textit{policy} by conditioning on past observations to predict future actions, or as a \textit{world action model} by additionally generating future states and observations. It can also serve as an \textit{inverse dynamics model}, predicting actions from observations. 

\noindent\textbf{Practical Flexibility in Deployment.}
Beyond these roles, the same mechanism enables practical flexibility, such as varying the history length to suit task requirements or masking out unavailable sensor modalities while still producing coherent trajectories. For example, when deploying the model on a robot without a force sensor, the force modality can either be treated as a prediction target in forceful tasks or discarded entirely in others. 

\noindent\textbf{Fine-grained Anomaly Detection.} MDF also enables localized likelihood estimation by injecting noise selectively into specific timesteps and modalities, rather than globally corrupting the entire trajectory. This design makes it possible not only to detect anomalies but also to precisely identify their source.

Diffusion models are trained to maximize a variational lower bound of the data log-likelihood (Eq.~\ref{eq:diffusion_train_obj}), and have therefore been applied to likelihood-based tasks such as density estimation~\cite{kingma2021variational} and anomaly detection~\cite{xue2025reactive, jianganomalies}. In this setting, anomalous samples are those with low likelihood under the learned distribution. Given a trajectory $\tau$, its likelihood can be estimated by \emph{measuring how well the diffusion model can recover $\tau$ after it has been corrupted by $k$ steps of the forward diffusion process}~\cite{xue2025reactive}. Concretely, Gaussian noise is added to $\tau$ through the forward process $q(\tau^k)$, after which we compare the posterior $q(\tau^{k-1} \mid \tau^k, \tau^0)$ with the learned reverse process $p_\theta(\tau^{k-1} \mid \tau^k)$. We estimate the anomaly score $D$ over a set of noise levels $I$ as
\begin{equation}
\vspace{-1mm}
\label{eq:likelihood}
D(I) = \frac{1}{|I|} \sum_{i \in I} D_{\mathrm{KL}} \left( q(\mathbf{\tau}^{i-1} \mid \mathbf{\tau}^i, \mathbf{\tau}^0) | p_\theta(\mathbf{\tau}^{i-1} \mid \mathbf{\tau}^i) \right)
\end{equation}

Building on this localized corruption mechanism, we design a fine-grained anomaly detection algorithm (Alg.~\ref{alg:anomaly_detection}) based on modality-time sweeping. Instead of perturbing the entire trajectory, our approach selectively injects noise into individual entries, each defined by a specific timestep and modality. By measuring how much each localized perturbation deviates from the model’s expected behavior, the method can not only detect the presence of anomalies but also pinpoint their precise timestep and modality. For example, abnormal point cloud data may indicate a faulty or obstructed camera, while abnormal force readings likely suggest that the robot is experiencing an external disturbance

\subsection{Training and implementation details}
\label{method:training}

We first pretrain the point cloud tokenizers~\cite{luo2021diffusion} separately on partial and full point clouds for 50k steps using a batch size of 512 and the Adam optimizer~\cite{kingma2014adam}. During MDF training, the pretrained tokenizers are frozen. The sequence length is set to 10, though both the history length and prediction horizon can be flexibly adjusted at test time.

As illustrated in Fig.~\ref{fig:system}, we randomly sample a 2D noise-level matrix during training. We adopt the Square Cosine Schedule~\cite{nichol2021improved} with 1000 denoising steps for training. At inference time, we perform full-sequence denoising (sec.~\ref{method:inference}) with 200 steps using the DDIM sampler~\cite{song2020denoising}. To further accelerate inference, we leverage CUDA Graphs, enabling the model to run at 10 Hz. The model is trained for 1 million gradient steps with the Adam optimizer~\cite{kingma2014adam}, using a batch size of 1024 and a learning rate of $5 \times 10^{-4}$.

\begin{center}
\vspace{-5mm}
    \begin{algorithm}[h]
\caption{Anomaly Localization via Modality-Time Sweeping}
\label{alg:anomaly_detection}
\begin{algorithmic}[1]
\REQUIRE Trajectory $\mathbf{x}$, noise levels matrix $\mathbf{K}\in \{0,\ldots,K\}^{T \times M}$, masked diffusion model $p_\theta$

\FOR{each timestep $t$ and modality $m$}
    \STATE Initialize noise matrix $\mathbf{K} \leftarrow \mathbf{0}$
    \STATE Set $\mathbf{K}_{t,m} \leftarrow i$ 
    \STATE Sample corrupted trajectory $\tau^{\mathbf{K}} \sim q(\tau^{\mathbf{K}} |\tau^0)$
    \STATE Compute and store KL divergence: $\mathcal{D}_{t,m} \leftarrow \mathrm{KL}\left(q(\tau^{k-1} \mid \tau^k, \tau^0) \,\|\, p_\theta(\tau^{k-1} \mid \tau^k)\right)$
    \STATE Compute KL divergence using Eq.~\ref{eq:likelihood}
\ENDFOR
\STATE \Return $(t^*, m^*) = \arg\max_{t,m} \mathcal{D}_{t,m}$
\end{algorithmic}
\end{algorithm}
\vspace{-6mm}
\end{center}
\section{Experiments}
We evaluate our methods on 3 contact-rich manipulation tasks in simulation and 2 forceful manipulation tasks in real world.
Through the experiments, we seek to answer the following questions:
\begin{itemize}
    \item \textbf{Performance}. MDF learns multiple probability distributions simultaneously. Can we match or even outperform the task-specific architecture for policy learning and anomaly detection?
    \item \textbf{Robustness}. Compared to baselines, does MDF provide additional robustness to sensory noise?
    \item \textbf{Flexibility}. To what extent can MDF be reconfigured at test time in terms of history length and input modalities?
    
\end{itemize}

\subsection{Action Generation for Contact-rich Manipulation tasks}
We first evaluate the robustness of our model in generating robot actions when provided with only partial or noisy observations. Specifically, we test on three simulated contact-rich manipulation tasks adapted from Nvidia Factory~\cite{narang2022factory}. 

\textbf{Nut Thread} A KUKA iiwa robot equipped with a Robotiq 3F gripper is required to thread an M16 nut onto a fixed bolt. Success depends on aligning the nut within the thread tolerance before screwing. The primary difficulty stems from frequent occlusions of the nut by the rotating gripper.

\textbf{Gear Mesh} A Franka Emika Panda robot must insert a gear into a partially assembled gearbox. The gear must first be guided through a supporting peg, then simultaneously meshed with two neighboring gears of different sizes. 

\textbf{Peg Insert} The robot must insert a cylindrical peg into a tight-fitting hole. The clearance is small, requiring accurate alignment and controlled contact. 
\begin{figure}[t]
    \centering
    \vspace{2mm}
    \includegraphics[width=1\linewidth]{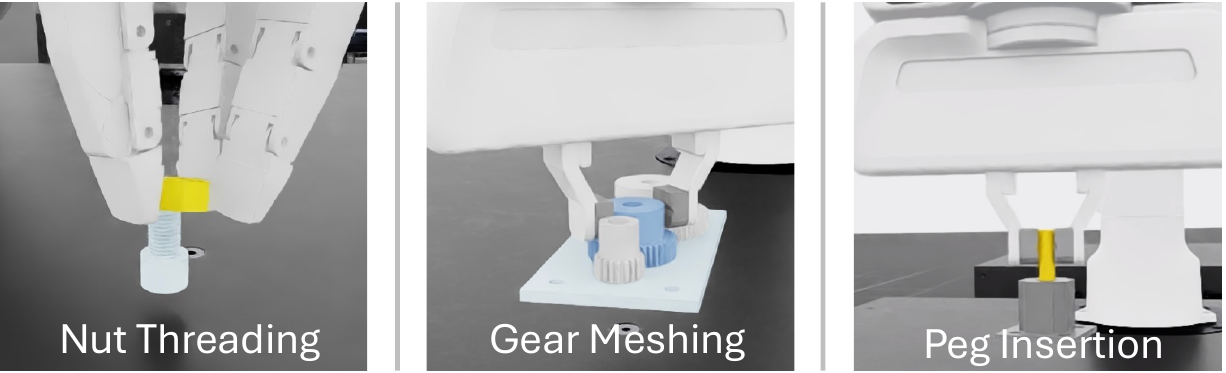}
    \caption{Contact-rich manipulation tasks in IsaacSim used in our experiments}
    \label{fig:sim-tasks}
    \vspace{-2mm}
\end{figure}
\subsubsection{Dataset collection} Teleoperation for contact-rich manipulation tasks in simulation is challenging due to the lack of feedback. We instead train a state-based RL policy using PPO to collect demonstrations paired with partial observations such as RGB image and point cloud. Collecting demonstrations with state-based RL for orbservation-based policy is a common technique in both online~\cite{kumar2021rma} and offline settings~\cite{chen2025clutterdexgrasp}. In this work, we focus on the offline supervised learning setting. We collect 10000 trajectories for Nut Thread and 7000 each for the Gear Mesh and Peg Insert.

\subsubsection{Baselines and ablations}
We select two state-of-the-art manipulation methods as baselines and ablate the design choice of our method.

\noindent\textbf{Unified World Model (UWM)}~\cite{zhu2025unified}. UWM trains a single model over video and action with independent noise levels. Similar to our method, UWM can also function as both a policy and a world model, but it lacks the capacity for geometric and force reasoning. Moreover, UWM has a fixed history length that cannot be adjusted at test time.
 
\noindent\textbf{3D Diffusion Policy (DP3)}~\cite{ze20243d}. DP3 incorporates 3D visual representations into diffusion policies~\cite{chi2023diffusion} with a specially designed point encoder. DP3 is an optimized architecture for 3D policy learning.

\noindent\textbf{MDF-Policy and MDF-WA}. At test time, MDF can be sampled in two modes for action generation. In the \textit{policy} mode, the model directly generates future actions. In the \textit{world action} (WA) mode, the model additionally predicts future observations and states as consequences of the actions (see Fig.~\ref{fig:noise-schedule-combined}). In both modes, the model also estimates the history state, i.e., the full point cloud, which serves as an aggregated representation of object pose and geometry. 

\noindent\textbf{Ablations}. We further examine the contributions of different components through ablation studies. Specifically, we evaluate \textit{MDF-No Wrench}, which removes force signals to test the importance of force reasoning, and \textit{MDF-Policy-No State Estimation}, where the history of full point cloud is not predicted during sampling.

\begin{table}
    \centering
    \begin{tabular}{@{}c|c|c|c@{}}
        \toprule
        Method & Nut Thread & Gear Mesh & Peg Insert \\ 
        \midrule
        DP3~\cite{ze20243d}            & 96\%  & 80\% & \textbf{84\%} \\
        UWM~\cite{zhu2025unified}      & 96\%  & 54\% & 58\% \\
        MDF-Policy                     & \textbf{100\%} & \textbf{86\%} & 80\% \\
        MDF-WA                    & 92\%  & 84\% & 78\% \\
        \midrule
        MDF-Policy-Noisy PC            & \textbf{94\%}  & \textbf{84\%} & \textbf{86\%}  \\
        DP3-Noisy PC                   & 78\%  & 68\% & 76\% \\
        \midrule
        MDF-policy-No wrench           & 72\% & 78\% & 74\%  \\
        MDF-Policy-No state estimation                      & 74\% & 74\% &  70\% \\
        \midrule
    \end{tabular}
\caption{We evaluate every method for 50 random configurations and report the success rate. }
\label{table:sim_result}
\vspace{-8mm}
\end{table}

\subsubsection{Results}
The success rates are reported in Table~\ref{table:sim_result}. 

\noindent\textbf{Is MDF able to match the performance of state-of-the-art policy for manipulation?}    Although MDF is trained jointly to model multiple probability distribution, it is on-par with or even outperforms specialized architectures such as DP3. For example, MDF-Policy achieves 100\% success rate in Nut Thread, outperforming DP3’s 96\%, and remains competitive on other benchmarks. Compared to the previous unified model, UWM, MDF consistently achieves higher success rates, which we attribute to its ability to perform both 3D geometric and force reasoning. Moreover, MDF can additionally leverage the full point cloud if available, and achieve a $100\%$ success rate on Peg Insert. While this setting provides privileged information, we argue that it remains realistic in semi-structured scenarios such as factories, where objects are known and specialized state estimation systems can be deployed to estimate poses reliably. The ability to accommodate different sensor setups is a major advantage over end-to-end policies with fixed inputs.

\noindent\textbf{Does MDF training improve the robustness of the model to sensor noise?}
To assess the robustness of the model, we injected random translations into the point cloud input to mimic camera calibration errors. Under this perturbation, MDF maintains strong performance, dropping only 4\% in Nut Thread and 2\% in Gear Mesh, while DP3 dropped by 18\% and 12\%. This is because during training, MDF is trained to denoise the partially corrupted input. 

\begin{figure}[t]
    \centering
    \includegraphics[width=0.9\linewidth]{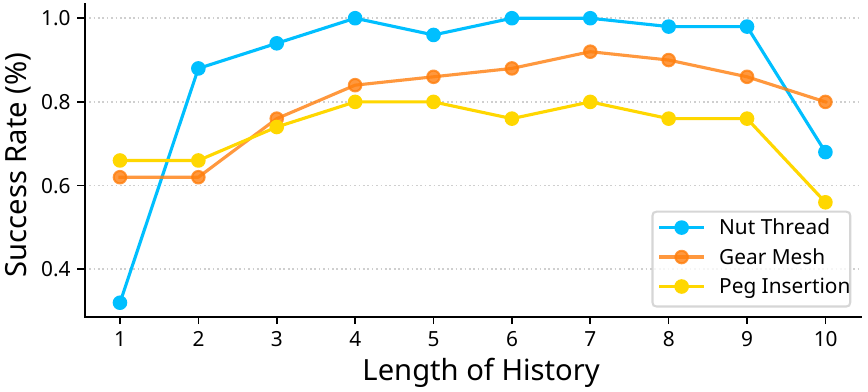}
    \vspace{-4pt}
    \caption{The history length of MDF can be adjusted dynamically at test time to accommodate task requirements.}
    \label{fig:mdf-history}
    \vspace{-6mm}
\end{figure}
\noindent\textbf{Can MDF handle different history lengths?} Unlike DP3 and UWM, which are trained with a fixed history length, the history length of MDF can be adapted at the test time based on the tasks requirements. The adaptibility to context length is a critical capability for large-scale multi-task learning. We demonstrate this capability of MDF in Fig.~\ref{fig:mdf-history}.

\noindent\textbf{Is dynamics modeling helpful for action generation at test time?}
We also observe that MDF-WA, which incorporates explicit dynamics modeling into action generation, performs worse than MDF-Policy. We hypothesize that this is due to the relatively short-horizon nature of the evaluated tasks, which do not require complex, system-2–style reasoning.  
\begin{figure}[b]
    \centering
    \vspace{-5mm}
    \includegraphics[width=1\linewidth]{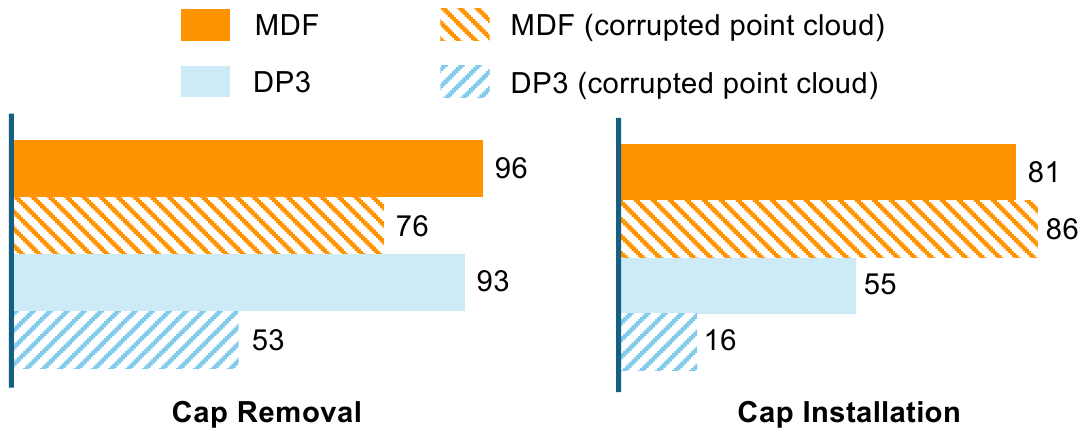}
    \vspace{-5mm}
    \caption{MDF demonstrates better robustness to noisy sensory input.}
    \label{fig:real_results}
\end{figure}

\noindent\textbf{Force reasoning and state estimation are important for contact-rich tasks}. Removing the force input or restricting the model to sampling future actions without estimating history states leads to performance drops of 6\%–28\%.
\begin{figure*}[ht]
    \centering
    \includegraphics[width=\linewidth]{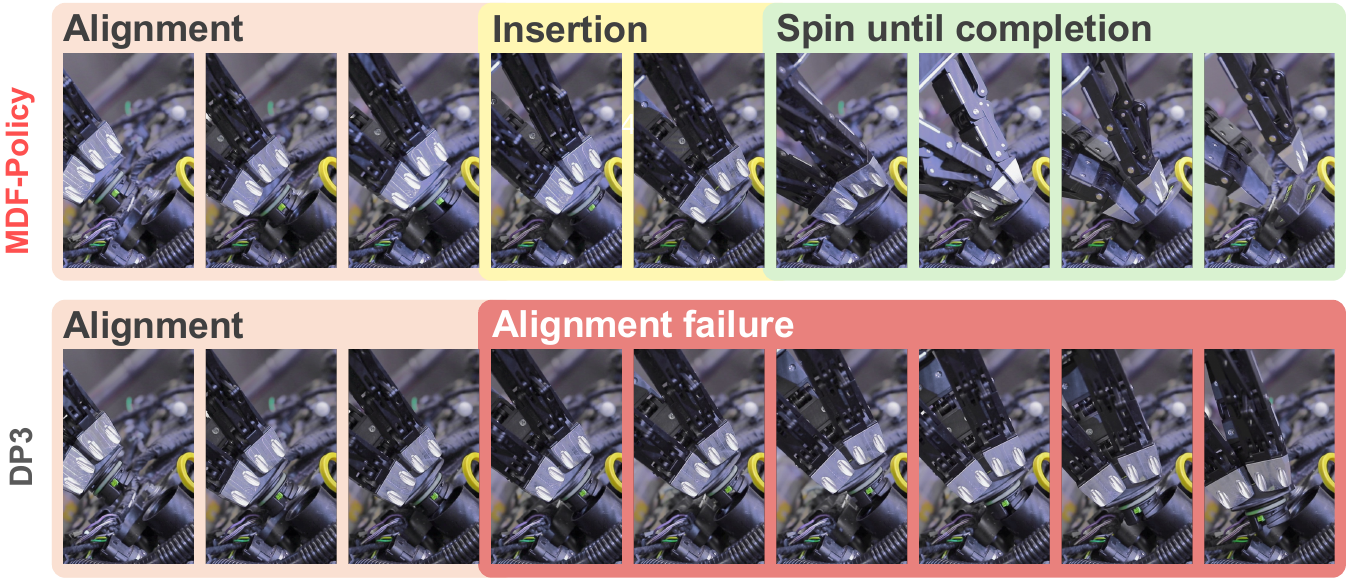}
    \caption{We compare MDF with DP3 on two real-world forceful manipulation tasks. For each task, we compute the score over 20 trials (160 in total), the grading standards can be found in Section \ref{sec:real-world}. MDF's noise-as-masking training scheme allows it to be more robust to noisy observations.}
    \label{fig:cap_rollout}
    \vspace{-5mm}
\end{figure*}

\subsection{Anomaly Localizaiton}
To evaluate the ability of our model to localize anomalous events, we design a fine-grained anomaly detection benchmark. Given a multimodal sequence, the goal is to precisely identify the timestep and modality in which anomaly occurs. To simulate anomalies, we corrupt point cloud observations by injecting random points and perturb wrench measurements with additive noise, mimicking external disturbances such as unexpected pushes. Each multimodal sequence is 10-steps long and contains 6 modalities, so a random guess will result in 1.67\% accuracy.
\begin{table}[t]
    \centering
    \begin{tabular}{@{}c|cc|cc@{}}
        \toprule
        \multirow{2}{*}{Method} & \multicolumn{2}{c|}{Wrench} & \multicolumn{2}{c}{Point cloud} \\ 
        \cmidrule(lr){2-3} \cmidrule(lr){4-5}
         & Time & Time-Mod & Time & Time-Mod \\
        \midrule
        MDF-sweeping & \textbf{73.8\%} & \textbf{66.0\%}& \textbf{100\%} & \textbf{77.7\%} \\
        MDF-global   & 65.6\% & 48.9\% & \textbf{100\%} & 63.6\% \\
        ImDiffusion  & 52.73\% & 3.52\% & 99.61\% & 5.47\%  \\
        \bottomrule
    \end{tabular}
    \caption{Results for anomaly localization}
    \label{table:anomaly_result}
    \vspace{-5mm}
\end{table}

Our proposed method, MDF-sweeping, localizes anomalies by selectively perturbing individual entries(Table~\ref{fig:noise-schedule-combined}). We compare against two baselines:
\begin{itemize}
    \item \textbf{MDF-global} applies a global noise level uniformly across the trajectory (setting the same values for entire noise level matrix) and computes entry-wise likelihoods, but suffers from reduced accuracy since the corrupted context contaminates the reference distribution.
    \item \textbf{ImDiffusion}~\cite{chen2023imdiffusion}, a state-of-the-art approach that masks out time-series entries and imputes them with a diffusion model, using imputation errors as anomaly scores.
\end{itemize}

Table~\ref{table:anomaly_result} summarizes the results. ImDiffusion achieves reasonable accuracy in identifying anomalous timesteps but fails to pinpoint the modality, as reflected by its sharp drop in the Time-Mod scores. MDF-global performs better but remains limited by global corruption. In contrast, MDF-sweeping achieves the highest localization accuracy across both wrench and point cloud modalities, particularly excelling in the challenging modality-time localization setting.

\subsection{Real-World Car Maintenance Manipulation Tasks} \label{sec:real-world}

We further evaluate our method on two challenging real-world car-maintenance tasks on a real vehicle engine with a 7 DoF Kuka LBR iiwa arm and Robotiq 3F gripper. 

\textbf{Oil Cap Installation} A KUKA robot is required to install an oil cap in the oil filler port. We grade this task with three partial-credit checkpoints: alignment (0.5 pt), insertion and rotation (0.75 pt), and cap fully locked (1.0 pt).



\textbf{Oil Cap Removal} A KUKA robot is required to completely remove a locked oil cap from the filler port. Similar to the Installation task, we grade this task with four partial-credit checkpoints: alignment (0.25pt), grip (0.5pt), rotation to cap unlocked state (0.75pt), and removal (1.0pt).

\subsubsection{Dataset Collection}
Both Oil Cap Installation and Removal demonstration datasets are collected through teleoperation. Point cloud are captured by a Zivid 2 camera.

\subsubsection{Results}
We compare our model with DP3 on both tasks. As shown in Fig.~\ref{fig:real_results}, MDF achieves up to a $26\%$ higher success rate. In the Oil Cap Installation task, 3D Diffusion Policy often stops turning prematurely and loosens its grip before the cap is fully locked. We conjecture that the absence of force input prevents DP3 from accurately reasoning about the cap’s locking state.

\textbf{Robustness Against Corrupted Input}.
Further, we demonstrate MDF's robustness against noisy input. Specifically, we adopt an alternative camera profile with a shorter capture time, producing noisier point clouds with increased missing regions. MDF's ability to dynamically define input noise level on a specific modality excels in this scenario, outperforming DP3 by 23\% and 70\% respectively (Fig.~\ref{fig:real_results}). 

An example rollout of the Oil Cap Installation task is presented in Figure \ref{fig:cap_rollout}. With identically corrupted point cloud perception, DP3 mis-aligns with the oil filler port and catastrophically fails to recover. Meanwhile, MDF is robust against noisy perception and successfully completes all three stages of this task. 
\begin{figure}[t]
    \centering
    \includegraphics[width=1\linewidth]{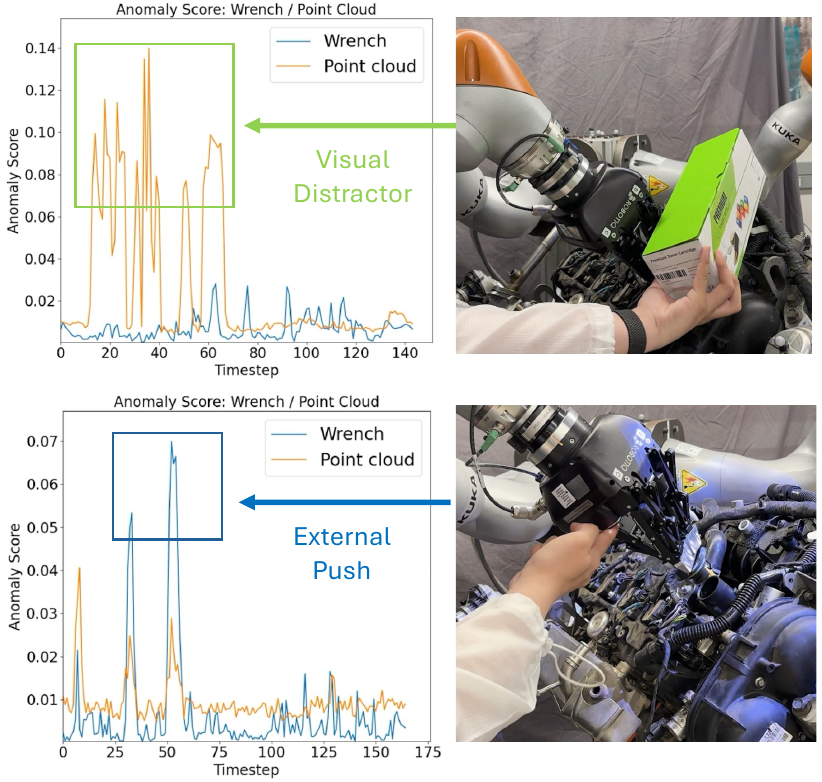}
    \vspace{-12pt}
    \caption{MDF enables fine-grained, per-modality anomaly detection.}
    \label{fig:real_anomaly}
    \vspace{-6mm}
\end{figure}

\textbf{Real-time Fine-grained Anomaly Detection}.
MDF enables fine-grained, per-modality anomaly detection at each timestep using Eq.~\ref{eq:likelihood}. As illustrated in Fig.~\ref{fig:real_anomaly}, the anomaly score rises selectively depending on the disturbance type. A visual distractor primarily increases the point cloud anomaly score while leaving the wrench modality largely unaffected. Conversely, an external physical push produces a pronounced spike in the wrench anomaly score, with minimal change in the score of point cloud. These results demonstrate MDF’s ability to localize anomalies across modalities.





\section{Conclusion}
\label{sec:conclusion}
We present Multimodal Diffusion Forcing (MDF), a unified framework for multimodal sequence modeling that captures the interplay of different modalities over time. MDF introduces a 2D \emph{Modality-Time Noise Level Matrix} that enables fine-grained control over the sampling distribution. At test time, MDF demonstrates strong 3D and force reasoning. The framework is highly flexible, supporting variable sensory modalities and adaptable history lengths, and robust to noisy observations. Furthermore, we showcase its versatility through applications such as fine-grained anomaly detection.

\textbf{Limitation and future work}: 1. Training efficiency: MDF jointly learns to capture many distributions, which poses a challenging optimization problem. A more targeted training strategy could improve efficiency by focusing on the distributions most relevant to downstream tasks. 2. Heterogeneous training: MDF naturally supports learning from heterogeneous datasets containing different subsets of input modalities. Scaling this direction further could enhance generalization and improve performance across diverse multimodal tasks.








\bibliographystyle{IEEEtran}
\bibliography{example}

\end{document}